# Recommendations for Marketing Campaigns in Telecommunication Business based on the footprint analysis

J. Sidorova[1], L.Skold[2], O. Rosander[1], L. Lundberg[1]

[1]Blekinge Institute of Technology
[2]Telenor
julia.a.sidorova@gmail.com
lars.lundberg@bth.se

**Abstract.** A major investment made by a telecom operator goes into the infrastructure and its maintenance, while business revenues are proportional to how big and good the customer base is. We present a data-driven analytic strategy based on combinatorial optimization and analysis of historical data. The data cover historical mobility of the users in one region of Sweden during a week. Applying the proposed method to the case study, we have identified the optimal proportion of geo-demographic segments in the customer base, developed a functionality to assess the potential of a planned marketing campaign, and explored the problem of an optimal number and types of the geo-demographic segments to target through marketing campaigns. With the help of fuzzy logic, the conclusions of data analysis are automatically translated into comprehensible recommendations in a natural language.

## 1 Introduction

In the telecommunication industry, the lion's share of capital is spent on the infrastructure and its maintenance. The revenues are dependent on the size and the quality of the customer base: without a customer there is no business, yet satisfying everyone is simply not feasible, unless the right ones are chosen and others are let to go [1]. The population in Sweden is growing rapidly due to immigration. In this light, the issue of infrastructure upgrades to provide telecommunication services is of importance. New antennas can be installed at hot spots of user demand, which will require an investment, and/or the clientele expansion can be carried out in a planned manner to promote the exploitation of the infrastructure in the less loaded geographical zones. In this paper, we explore the second alternative and formulate the recommendations with respect to an intelligent expansion of the customerbase. Specifically, the problem we concern ourselves with is how to find a balanced user portfolio in order to optimally exploit the infrastructure and get maximum benefit from the investments. The intuitive observation, which motivated our solution, is as follows. The individual mobility patterns of different user segments sum up into a collective footprint, which the whole customer base produces on the infrastructure in a time-continuous manner. The desired property of such a collective footprint is that it does not exhibit skinny peaks and gaps in time. The closer to the optimal "heavy and yet even load" scenario, the better the infrastructure is exploited.

Methodology-wise, the literature in telecommunications research is abundant with optimization approaches formulated for the exploitation of telecommunication networks under the disguise of the problems (at first glance possibly unrelated to ours) such as optimal location of cell towers, optimization of base stations deployment and so on, e.g. [2]-[5]. For example, the dual formulation of the optimal positioning of new cell towers turns out to be our problem of finding an optimal portfolio with user segments [6]. Thus, the literature suggests a methodological appropriateness of a linear programming formulation. The research gap we have noticed and fill in is that, to our best knowledge, such works do not make use of historical data.

Another question is how to represent different user groups within the linear programming system. To this end, postcode-based geo-demographic segments are both strong predictors of user behavior and operational user handles in marketing campaigns. Compared with conventional occupational measures of social class, postcode classifications typically achieve higher levels of discrimination, whether averaged across a random basket of behaviors recorded on the Target Group Index or surveys of citizen satisfaction with the provision of local authority services. One of the reasons that segmentation systems like MOSAIC are so effective is that they are created by combining statistical averages for both census data and consumer spending data in pre-defined geographical units [7]. The postcode descriptors allow us powerful means to unravel lifestyle differences in ways that are difficult to distinguish using conventional survey research given limited sources and sample size constraints [8]. For example, it was demonstrated that middle-class MOSAIC categories in the UK such as 'New Urban Colonists', 'Bungalow Retirement', 'Gentrified Villages' and 'Conservative Values', whilst very similar in terms of overall social status, nonetheless register widely different public attitudes and voting intentions, show support for different kinds of charities and preferences for different media as well as different forms of consumption. Geodemographic categories correlate to diabetes propensity [9], school students' performance [8], broadband access and availability [7], and so on. Industries rely increasingly on geodemographic segmentation to classify their markets when acquiring new customers [10]. The localized versions of MOSAIC have been developed for a number of countries, including the USA and the EU countries. The main geodemographic systems are in competition with each other and the exact details of the data and methods for generating lifestyles segments are never released [11] and, as a result, the specific variables or the derivations of these variables are unknown. Faced with this uncertainty, we have tried two different geo-demographic segmenations and diverse levels of granularity in those.

The contributions of this paper are as follows. A data-driven methodological framework has been formulated as a classical resource allocation problem to calculate the degree of desirability of different groups of clients based on the footprint they and the rest of the population produce on the infrastructure. We have verified our idea in a case study: the optimal proportion of geo-demographic segments in the customer base was identified. As a natural consequence of the approach, a functionality has been developed to assess the potential of a planned marketing campaign. Then, the problem of finding the optimal number of geo-demographic segments to target simultaneously has been addressed empirically. We compare different marketing scenarios for the two segmentations available: the MOSAIC with two levels of granularity (15 segments and 46 subsegments) and six segments developed by InsightOne specifically for Telenor. We investigate the trade off: more effort is required to target a greater number of segments, but a finer discrimination would lead to a better control of infrastructure exploitation and, thus, higher revenues. Fuzzy logic modeling is used to build an interface between a manager and big data processing to translate the conclusions into a comprehensible summary in a natural language. While translating numeric answers into recommendations, insignificant numeric deviations are gotten rid of due to a formulation with qualitatively different hedges: extremely, very, rather, and hardly. The following queries have been formulated on the mobility database:

- Which segments are extremely/very/rather desired?
- How the infrastructure is currently exploited: extremely/very/rather or hardly efficiently?
- If the identified segments are boosted as expected as a result of a corresponding marketing camapign, how will the exploitation efficiency change?

The rest of the paper is organized as follows. In Section 2, the data set is described. Section 3 explains the classical resource allocation formulation for our problem with a data-driven aspect. Section 4 addresses the problem of building an interface between a manager and big data processing. In Section 5, experiments are covered, and the trade-off between the granularity of segmentation and the impact of the resulting footprint on the infrastructure is investigated. Finally, the conclusions are drawn in Section 6.

## 2   Geospatial and geo-demographic data

The study has been conducted on anonymized geospatial and geo-demographic data provided by a Scandinavian telecommunication operator. The data consist of CDRs (Call Detail Records) containing historical location data and calls made during one week in a midsized region in Sweden with more than one thousand radio cells. Several cells can be located on the same antenna. The cell density varies in different areas and is higher in city centers, compared to rural areas. The locations of 27010 clients are registered together with which cell serves the client. The location is registered every five minutes. In the periods when the client does not generate any traffic, she does not make any impact on the infrastructure and such periods of inactivity are not included in the resource allocation analysis. Every client in the database is labeled with her geo-demographic segment. The fields of the database used in this study are:

- the cells IDs with the information about which a user it served at different time points,
- the location coordinates of the cells,
- the time stamps of every event with a 5 minute resolution,
- the MOSAIC geo-demographic segment (and subsegment) for each client, and
- the Telenor geo-demographic segment for each client.

There are 15 (and 46) MOSAIC segments (and subsegments) present in the geographic region under analysis; for their detailed description the reader is referred to [12]. The six in-house segments were developed by Telenor in collaboration with

InsightOne, and, to our best knowledge, though not conceptually different from MOSAIC, they are especially crafted for marketing in telecommunication businesses.

## 3  The Combinatorial Optimization Module

The individual mobility patterns of different user segments sum up into their collective footprint, which the whole customer base produces on the infrastructure in a time-continuous manner. A desired property of such a collective footprint is that it does not exhibit skinny peaks and gaps in time. The closer to the optimal "even load" scenario, the better the infrastructure is exploited. The model's variables are the following.

*Variables:*
- `clientSet`: set of with IDs of clients;
- `I`: the set with geo-demographic segments {$segment_1$, …, $segment_k$};
- `D`: the mobility data for a region that for each user contain client's ID, client's geo-demographic segment, time stamps when the client generated traffic, and which antenna served the client.
- $S_i$: the number of subscribers that belong to a geo-demographic segment $i$;
- $S_i^*$: the optimal number of subscribers that belong to a geo-demographic segment $i$;
- $S_{i,t,j}$ : the footprint by segment $i$, i.e. the number of subscribers that belong to a geo-demographic segment $i$, at time moment $t$, who are registered with a particular cell $j$;
- $C_j$: the capacity of cell $j$ in terms of how many persons it can safely handle simultaneously;
- **x**: the vector with the scaling coefficients for the geo-demographic segments or other groups such as IS clients;
- $x_{IS}$: the coeffcient for the IS segment from the vector **x**;
- $N_{t,j}$: number of users at cell j at time t.

The problem of finding an optimal combination of user segments, given that we want to maximize the overall number of users, who consume finite resources, belongs to a family of resource allocation problems. The formulation of our problem is as follows:

- *The vector **x** with the decision variables*

$$\mathbf{x} = \{ x_{CC}, x_{CA}, x_{MJM}, x_{QA}, x_T, x_{VA} \}.$$

The decision variables represent the scaling coefficients for each geo-demographic segment. In case of Telenor segmentation they are: cost-aware (CA), modern John/Mary (MJM), quality aware (QA), traditional (T), value aware (VA), and corporate clients (CC). A scaling coefficient $x_i$ is greater than 1, if the number of clients of a given geo-demographic segment is desired to be increased. For example, for the category in the customer base that is to be doubled $x_i = 2$. Similarly, if $x_i < 1$ for a geo-demographic segment, it means that the number of clients is to be reduced. The $x_i = 0$ value indicates that the segment is absolutely unwanted in the clientele. By formulation **x** is non-negative.

- *The objective function* seeks to maximize the number of subscribers:

$$Maximize \quad \Sigma_{i \in \{ CC, CA, MJM, QA, T, VA\}} S_i x_i \quad (1)$$

- *The restrictions*

$$\text{for all } j,t, \; \Sigma_{i \in \{CC, CA, MJM, QA, T, VA\}} \; S_{i,t,j} x_i \leq C_j \quad (2)$$

represent the observed number of persons in each user group at a particular time and served by a particular cell multiplied by the scaling coefficient. This value is required not to exceed the capacity of the cell $C_j$ in terms of how many persons it can handle at a time. In other words the restriction says: if the historical number of users are scaled with a coefficient for their geo-demographic category, the cells should not be overloaded.

A consensus reached in the literature [13]-[15] is that the mobility pattern for the subscribers is predictable due to strong spatio-temporal regularity. The corollary is that the increase in the number of subscribers in a given segment with a factor x will result in an increase of the load generated by the segment with a factor x for each time and cell.

The LP model is solved for the input data $D$ and the set of segments $I$:

$$(x_I, \; max\_obj_{I,D}) = combinatorial\_optimization(D,I). \quad (3)$$

The output is the vector with the optimal scaling coefficients $x_I$ and the maximum value of the objective function.

Consider a small example with two cells, two subscriber segments and three time slots. The footprint values are shown in Table I. The total number of subscribers in segment 1 is 60, and the total number of subscribers in segment 2 is 40 ($\mathbf{s} = (60, 40)^T$). The capacity of both radio cells is 200, i.e., $\mathbf{c} = (200, 200)^T$. The optimization problem becomes:

$$\text{Maximize } 60x_1 + 40x_2.$$

The LP problem has 6 restrictions:

$$\text{for } t_1, \text{ cell 1: } 40x_1 \leq 200,$$

$$\text{for } t_1, \text{ cell 2: } 20x_1 + 20x_2 \leq 200,$$

$$\text{for } t_2, \text{ cell 1: } 40x_1 \leq 200,$$

$$\text{for } t_2, \text{ cell 2: } 40x_2 \leq 200,$$

$$\text{for } t_3, \text{ cell 1: } 25x_1 + 25x_2 \leq 200,$$

$$\text{for } t_3, \text{ cell 2: } 10x_1 + 20x_2 \leq 200,$$

$$x \geq 0.$$

That is, we have the following:

$$A \begin{pmatrix} 40 & 0 \\ 20 & 20 \\ 40 & 0 \\ 0 & 40 \\ 25 & 25 \\ 10 & 20 \end{pmatrix}, \; Bc \begin{pmatrix} 10 \\ 01 \\ 10 \\ 01 \\ 10 \\ 01 \end{pmatrix} \begin{pmatrix} 200 \\ 200 \end{pmatrix} \begin{pmatrix} 200 \\ 200 \\ 200 \\ 200 \\ 200 \\ 200 \end{pmatrix}$$

Solving this LP problem yields the optimal $\mathbf{x} = (5, 3)^T$, corresponding to $\mathbf{s}^T\mathbf{x}$ of 420.

| Time slot | Cell 1 | | Cell 2 | |
|---|---|---|---|---|
| | Segment 1 | Segment 2 | Segment 1 | Segment 2 |
| $t_1$ | 40 | 0 | 20 | 20 |
| $t_2$ | 40 | 0 | 0 | 40 |
| $t_3$ | 25 | 25 | 10 | 20 |

**Table 1:** The number of subscribers in each segment for all time slots and cells for the small example.

Before we continue, we need to discuss some implicitly made assumptions that may be not necessarily fair. Firstly, *all the clients generate the same revenue*. Concrete tariffs are integrated in the form of coefficients of the objective function. Let the tariff for the user category $i$ be denoted with $R_i$. Then, the initial objective function from Equation 1 is extended into

$$\text{Maximize} \quad \Sigma_{i \in \{CC, CA, MJM, QA, T, VA\}} R_i S_i x_i.$$

Secondly, *the impact on the network produced by different users is the same.* The calculation of the impact on the network can be refined taking into account the historical traffic. Let the traffic generated by the user group be $T_i$. The restrictions from Equation 2 are modified:

$$\Sigma_{i \in \{CC, CA, MJM, QA, T, VA\}} T_{i,t,j} x_i \leq C_j.$$

these clarifications can be easily accommodated in the system, but currently the relevant knowledge about $R_i$ and $T_i$ is out of our reach.

## 4. Manager – Fuzzy Logic – Processing Big Data

In the era of big data a mapping is desired from multitudes of numeric data to its useful summary in a natural language with insignificant numeric deviations gotten rid of [16].

### 4.1 Notation and Definitions

**Definition** (in the style of [17]). A fuzzy set $A$ in $X$ is characterized by a membership function $f_A(x)$, which associates with each point in $X$ a real number in the interval $[0, 1]$, with the value of $f_A(x)$ at $x$ representing the "grade of membership" of $x$ in $A$. For the opposite quality: $f_{notA}(x) = 1 - f_A(x)$.

Fuzzy membership scores reflect the varying degree to which different cases belong to a set:

- Under the six value fuzzy set, there are six tiers of membership *1*: fully in, *0.9*: mostly but not fully in, *0.6*: more or less in, *0.4*: more or less out, *0.1*: mostly but not fully out, *0:* fully out.
- Thus, fuzzy sets combine qualitative and quantitative assessment: 1 and 0 are qualitative assignments ("fully in" and "fully out", respectively); values between 0 and 1 indicate partial membership. The 0.5 score is also qualitatively anchored, for it indicates the point of maximum ambiguity (fuzziness) in the assessment of whether a case is more "in" or "out" of a set.

For a comprehensive guide of good practices in fuzzy logic analysis in social sciences the reader is referred to, for example, [18].

**Linguistic hedges:**

- *Rather* will be added to a quality $A$, if the square root of its membership function $f_A(x)^{1/2}$ is close to *1*.

- *Very* will be added to a quality $A$, if the square of its membership function $f_A(x)^2$ is close to $1$.
- *Extremely* will be added to a quality $A$, if $f_A(x)^3$ is close to $1$.

The interpretation follows from the application of the hedge operator, which adds the quantifiers such as *very, rather, extremely*, to the membership function, for example $f_{veryA}(x) = f_A(x)^2$ [19]. Then, given the new membership function, the same principle applies: the closer to 1, the higher is the degree of membership. Inside a tier, the hedged membership functions obey an inclusion relation: *extremely f ⊂ very f ⊂ rather f*. As long as the same hedge applies to the value of the membership function, numeric differences are held as insignificant (they are quantitative), once the condition for a new hedge is met, the situation changes in a qualitative way, e.g. comparing two attempts the Collective Ear can confirm that both of those are very representative of anger, and thus the numeric differences between their assessments do not matter.

### 4.2. Interface Construction:

**Desirability of user groups:** in order to make the vector $x$ express the desirability of different user groups, the scaling coefficients are normalized so that the largest normalized coefficient is equal to $1$. Then, $x_i$ is naturally interpreted to be the $f_{desired}(segment\ i)$, and the tests for different hedges are applicable (See Section 5.1).

**Assessing the success of a marketing campaign** and the consequences of the modifications in the customer base can be simulated. The red line (see Figure 1) is the limit of success in the infrastructure exploitation and represents the most successful marketing campaign possible, i.e. with the membership value for efficiency equal to 1. The blue line has been calculated for the present $x$ is the starting point for a new marketing campaign and serves as a separation point between business expansion and losses, i.e. with the membership function equal to $0.5$. All the reasonable scenarios fall between the present and the best possible value.

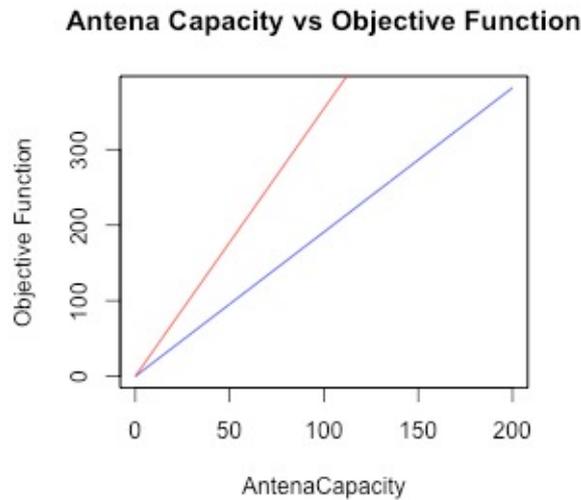

**Figure 1:** The maximum number of clients that can be served, given the current (blue)

and the optimized proportion (red) of segments in the customer base.

## Experiments

### 6.1 Optimal proportion of geo-demographic segments

Once the LP model was built from the data and solved with the gurobi solver [20] using HPI Future SoC Lab Hardware Resources, the vector with the optimal scaling coefficients was returned:

$$x = (0, 0.13, 0, 1.45, 4.85, 0.92)^T.$$

With the optimal proportion of the geo-demographic segments in the customer base, *57%* more clients can be safely served without inducing any upgrade costs or overloading the cells. The ideal coefficients are unlikely to be attained ever, but even moving towards the optimal proportion of the geo-demographic segments means moving towards higher revenues and more efficient antenna exploitation. In Figure 1, where the objective function (i.e. how many users can be provided service in a safe manner) is plotted against the cell capacity, the cell capacity has been taken to be an interval from 0 to 200 clients rather than an integer value, because the actual footprint depends on how the users are consuming services, as for example, streaming video (and creating a heavy footprint) versus just receiving an SMS (a light footprint).

Comparing the obtained values for $\{x_{CC}, x_{CA}, x_{MJM}, x_{QA}, x_T, x_{VA}\}$ to the coefficients in the current proportion, a conclusion is made whether the geo-demographic segment needs to be reduced or boosted. The recommended action is to boost the *Quality-Aware* and *Traditional* segments, get rid of *Corporate Clients* and *Modern John Mary* and partially reduce all the other segments. Vector *x* in the normalized form is:

$$x_{normalized} = (0, 0.02, 0, 0.29, 1, 0.18).$$

**Query 1:** Retrieve *very* desired segments.
```
for segment in I do{
      v_desired(segment)=FALSE
      IF (f_desired(segment)² ≥ 0.9) THEN v_desired(segment)=TRUE
}
```
In our case study, only the *Traditional* segment is very desired. Analogous tests on the value of the membership function are executed to check the applicability of other hedges.

If the old clientele is decided to be kept, additional restrictions are to be added to the LP: $x_i \geq 1$. Obviously, pleasing everyone has a negative effect on the slope of benefit generation. Figure 2 reflects the cost in terms of the objective value. As follows from the graphics, the minimum cell capacity of *165* is required to be able to reliably provide service to everyone and there is no additional revenue. From the capacity of *165* to *310* the benefit generation is slowed down by having to keep the old clients. Once the cell would serve more than *310* persons at a time, the restrictions $x_i \geq 1$ stop being tight, and keeping the old clients would not imply any potential losses, but currently the cell capacity is limited to 200 clients.

**Query 2:** How the infrastructure is currently exploited: extremely, very, rather or hardly efficiently?

$$f_{\text{efficently exploited}} = current\_obj \ (max\_obj)^{-1},$$

where *current_obj* is the maximum number of persons that the infrastructure can serve, given that the present proportion of the segments is kept (a linear, indiscriminative expansion of the customer base), and *max_obj* is the theoretically largest possible number of clients that can be served given the ideal proportions of the segments. Further the tests for the applicability of the linguistic hedges can be tried, as was explained in Section 4.1.

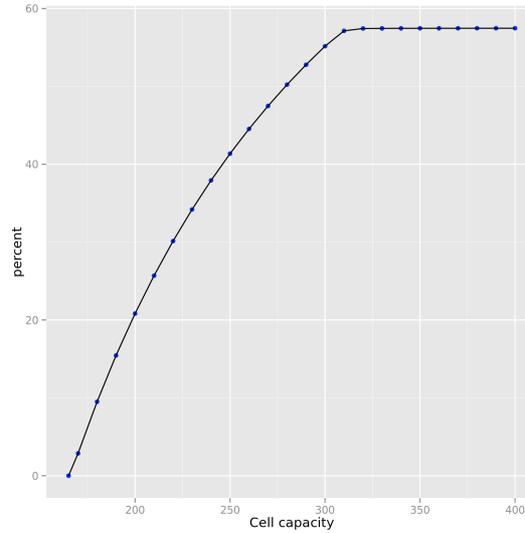

**Figure 2**: Keeping and pleasing every client slows down the percent of clients that can be served up to a critical value of cell capacity.

Once the optimal proportion of segments in the clientele (S) is known, it is direct to assess the optimality of the present customer base and changes in it:

$$f_{optimal}(S) = \Sigma_{all \ i} \ S_i \ (\Sigma_{all \ i} \ S^*_i)^{-1} \text{ and}$$
$$f_{optimal}(S^{new}) = \Sigma_{all \ i} \ S_i^{new} \ (\Sigma_{all \ i} \ S^*_i)^{-1}.$$

Suppose a marketing campaign *A* is expected to transform the current clientele $S_i$ into $S_i^{new}$:

$$\text{Action } A: S_i \rightarrow S_i^{new}.$$

Firstly, the LP is checked for feasibility, given $S^{new}$, i.e. the modifications must not violate any of the restrictions. The measure for efficiency for the potential of *A* is defined as:

$$f_{efficent}(A) = (f_{efficient}(S^{new}) - f_{efficient}(S))(1 - f_{efficient}(S))^{-1}.$$

**Query 3:** If in cause in a particular marketing campaign *A* the identified segments are boosted as expected, how will the exploitation efficiency change?

For example, within *A* the segments *Traditional* and *Quality-Aware* are planned to be boosted by *5-7%*. An alternative *B* implies boosting *Quality-Aware* and *Traditional* by *8-10%* and *1-3%,* respectively. According to their simulation details in Table 2. action *A* is *0.035* better than action *B*, but they fall into the same tier with respect to their potential. It does not make a qualitative difference which action to undertake.

| Corporate clients | 139 subscribers in the database |
|---|---|
| Cost aware | 4003 subscribers in the database |
| Modern John/Mary | 5963 subscribers in the database |
| Quality aware | 5805 subscribers in the database |
| Traditional | 6007 subscribers in the database |
| Value aware | 5093 subscribers in the database |
| New clients with A | [590, 826] |
| Expansion as a result of A feasible | yes |
| New clients with B | [524, 2382] |
| Expansion as a result of B feasible | yes |
| $f_{efficient\ with\ A}(S^{new})$ | [0.63, 0.65] |
| $f_{efficient\ with\ B}(S^{new})$ | [0.64, 0.68] |
| $f_{efficient}(S)$ | 0.63 |
| $f_{efficent}(A)$ | [0; 0.65] |
| $f_{efficent}(B)$ | [0.01; 0.13] |
| $E(f_{efficent}(A))$ | 0.06 |
| $E(f_{efficent}(B))$ | 0.025 |
| Same tier? | Yes |
| Conclusions | No difference |

**Table 1:** Comparison of the expected effects of action A and B targeting on Telenor segmentation.

### 5.2 Granularity vs. Efficiency

The more decision variables, the more degrees of freedom the LP model has, and naturally the higher value of the objective function can be achieved. A comparison of the performance for different segmentations is presented in Figure 3: 46 MOSAIC sub-segments, 15 MOSAIC segments, and 6 Telenor segments. To study the effect of granularity, we employed a greedy merge heuristic algorithm relying on the following heuristic.

*Heuristic 1: "at one step, merge the two segments, for which the scaling coefficients are closest".*

The choice of the heuristic was motivated by the observation, that if the scaling coefficients are close, then the corresponding segments should receive the same encouraging or discouraging force, and, when possible, the marketing campaign can be run once and target both of the segments at once. As was expected, with the 45 MOSAIC sub-segments the best optimization result has been achieved, due to the highest degree of freedom in the LP model. In line with this observation, when the number of segments reaches the number of the MOSAIC sub-segments, they demonstrate a similar success rate. The Telenor segments perform the worst, which is explainable with the fact that MOSAIC is a successful predictor of distinct qualities (footprint as a result of the life-style), while the Telenor segmentation partition was crafted to facilitate handy segment definitions. The results exhibit two trade-offs.

Firstly, while planning campaigns, on one hand more effort is required to target a greater number of segments, but on the other hand a finer discrimination would lead to a more optimal infrastructure exploitation and higher revenues. Secondly, on one hand, the merged MOSAIC segments would improve the performance compared to the in-house segmentation, but on the other hand the latter is more convenient for the marketing department to work with.

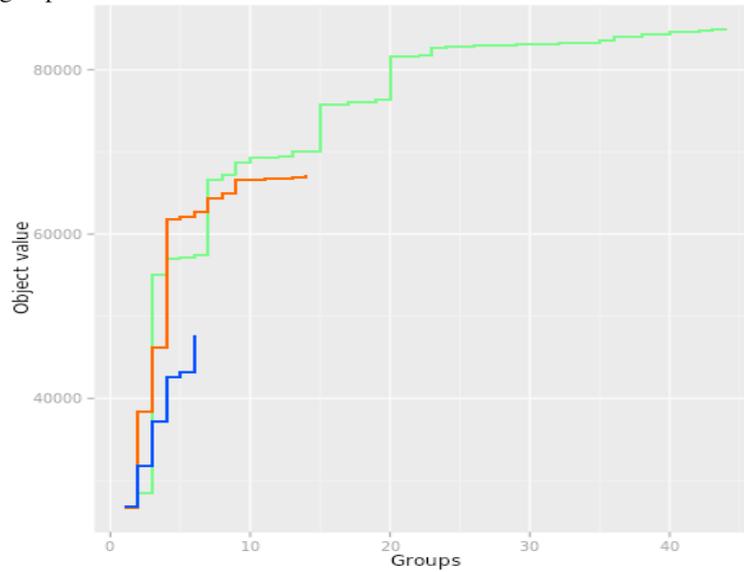

**Figure 3**: The objective value vs segment granularity: Telenor segments (blue), MOSAIC segments (red), and MOSAIC sub-segments (green).

## 6   Conclusions

This work focuses on how to grow the customer base of a telecommunications operator in an intelligent way which implies minimal additional expenditures on the infrastructure. Classical combinatorial optimization is behind the inner machinery of the proposed framework, and our contribution is the use of historical data. The numeric analytical results in this case study are translated into cognitively comprehensive conclusions unpolluted by insignificant numeric fluctuations. In the process of mining historical data from a telecommunication operator, the optimal proportion of geo-demographic segments in the customer base have been identified, a functionality was developed to assess the outcome of planned marketing campaigns, and the trade-off of the granularity of segmentation and the efficiency of campaigning has been explored. The above listed analytics have been implemented as queries over the database with historical mobility.

**Acknowledgments**